\documentclass[runningheads]{llncs}
\usepackage{cite}
\usepackage{amsmath,amssymb,amsfonts}
\usepackage{algorithm}
\usepackage{algorithmic}
\usepackage{graphicx}
\usepackage{textcomp}
\usepackage{multirow} 
\usepackage{makecell} 
\usepackage{booktabs}
\usepackage{bbm}
\usepackage{mathtools}
\usepackage{bbold}
\usepackage{caption}
\usepackage{marvosym}
\usepackage{ifsym}
\usepackage{xcolor}         
\newcommand{\ie}{\textit{i.e.}, }
\newcommand{\eg}{\textit{e.g.}, }
\newcommand{\etal}{\textit{et al. }}

\newcommand{\wrt}{\textit{w.r.t. }}
\usepackage[colorlinks,
            linkcolor=red,
            anchorcolor=red,
            citecolor=blue,
            ]{hyperref}     

\begin{document}
\title{MetaLR: Meta-tuning of Learning Rates for Transfer Learning in Medical Imaging}
\titlerunning{MetaLR}
\authorrunning{Y. Chen, L. Liu, et al.}
\author{Yixiong Chen\inst{1,2} 
\and 
Li Liu\inst{3}$^{(\textrm{\Letter})}$
\and
Jingxian Li\inst{4}
\and
Hua Jiang\inst{1,2}
\and \\
Chris Ding\inst{1}
\and
Zongwei Zhou\inst{5}
}

\institute{The Chinese University of Hong Kong (Shenzhen), China \and
Shenzhen Research Institute of Big Data, Shenzhen, China\\
\and The Hong Kong University of Science and Technology (Guangzhou), China
\email{avrillliu@hkust-gz.edu.cn}\\
\and Software School, Fudan University, Shanghai, China\\
\and Johns Hopkins University, Baltimore, USA}
\maketitle              
\begin{abstract}

In medical image analysis, transfer learning is a powerful method for deep neural networks (DNNs) to generalize well on limited medical data. Prior efforts have focused on developing pre-training algorithms on domains such as lung ultrasound, chest X-ray, and liver CT to bridge domain gaps. However, we find that model fine-tuning also plays a crucial role in adapting medical knowledge to target tasks. The common fine-tuning method is manually picking transferable layers (\eg the last few layers) to update, which is labor-expensive. In this work, we propose a meta-learning-based LR tuner, named MetaLR, to make different layers automatically co-adapt to downstream tasks based on their transferabilities across domains. MetaLR learns appropriate LRs for different layers in an online manner, preventing highly transferable layers from forgetting their medical representation abilities and driving less transferable layers to adapt actively to new domains. Extensive experiments on various medical applications show that MetaLR outperforms previous state-of-the-art (SOTA) fine-tuning strategies. \href{https://github.com/Schuture/MetaLR}{Codes} are released.

\keywords{Medical image analysis \and Meta-learning  \and Transfer learning.}
\end{abstract}
\section{Introduction}


Transfer learning has become a standard practice in medical image analysis as collecting and annotating data in clinical scenarios can be costly. The pre-trained parameters endow better generalization to DNNs than the models trained from scratch~\cite{tajbakhsh2016convolutional,chen2022generating}. A popular approach to enhancing model transferability is by pre-training on domains similar to the targets~\cite{zhou2019models,zhou2020comparing,riasatian2021fine,chen2021uscl,zhang2022hico}. However, utilizing specialized pre-training for all medical applications becomes impractical due to the diversity between domains and tasks and privacy concerns related to pre-training data.
Consequently, recent work~\cite{guo2019spottune,amiri2020fine,ro2021autolr,chambon2022improved} has focused on improving the generalization capabilities of existing pre-trained DNN backbones through fine-tuning techniques.

\begin{figure}[t]
\vspace{0cm}                          
\centering\centerline{\includegraphics[width=1.0\linewidth]{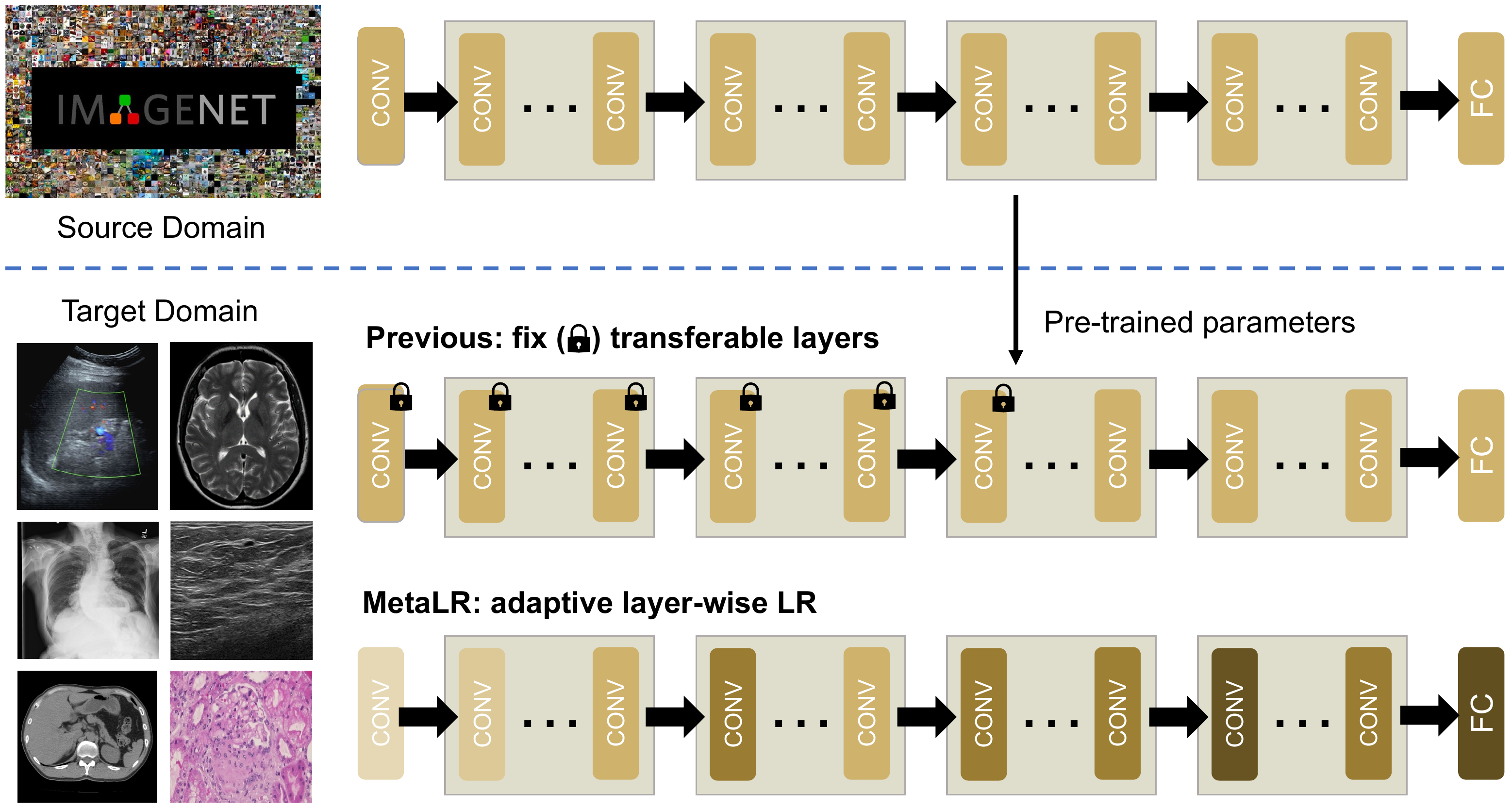}}
\caption{The motivation of MetaLR. Previous works fix transferable layers in pre-trained models to prevent them from catastrophic forgetting. It is inflexible and labor-expensive for this method to find the optimal scheme. MetaLR uses meta-learning to automatically optimize layer-wise LR for fine-tuning.}
\label{fig:motivation}
\vspace{-0.5cm}     
\end{figure}

Previous studies have shown that the transferability of shallower layers is often higher than that of deeper layers~\cite{yosinski2014transferable}. Layer-wise fine-tuning \cite{tajbakhsh2016convolutional}, which is a practical application of this finding, was introduced to preserve the transferable low-level knowledge. Moreover, recent studys in medical fine-tuning have revealed that the transferability of pre-trained knowledge significantly varies across different downstream datasets \cite{wang2018interactive} and layers \cite{vrbanvcivc2020transfer}. In some cases, transferability might even be irregular among layers for domains far from pre-training data \cite{chen2022rethinking}. \textit{Given the diverse medical domains and model architectures, there is currently no universal guideline to follow to determine whether a particular layer should be retrained for a given target domain.}

To search for optimal layer combinations for fine-tuning, manually selecting transferable layers \cite{tajbakhsh2016convolutional,amiri2020fine} can be a solution, but it requires a significant amount of human labor and computational cost. In order to address this issue and improve the flexibility of fine-tuning strategies, we propose controlling the fine-tuning process with layer-wise learning rates (LRs), rather than simply manually fixing or updating the layers (see Fig. \ref{fig:motivation}). Our proposed algorithm, Meta Learning Rate (MetaLR), is based on meta-learning~\cite{franceschi2018bilevel} and adaptively adjusts LRs for each layer according to transfer feedback. It treats the layer-wise LRs as meta-knowledge and optimizes them to improve the model generalization. Larger LRs indicate less transferability of corresponding layers and require more updating, while smaller LRs preserve transferable knowledge in the layers. Inspired by \cite{ren2018learning}, we use an online adaptation strategy of LRs with a time complexity of $O(n)$, instead of the computationally-expensive bi-level $O(n^2)$ meta-learning. We also enhance the algorithm's performance and stability with a proportional hyper-LR (LR for LR) and a validation scheme on training data batches.

In summary, this work makes the following three contributions. 1) We introduce MetaLR, a meta-learning-based LR tuner that can adaptively adjust layer-wise LRs based on transfer learning feedback from various medical domains. 2) We enhance MetaLR with a proportional hyper-LR and a validation scheme using batched training data to improve the algorithm's stability and efficacy. 3) Extensive experiments on both lesion detection and tumor segmentation tasks were conducted to demonstrate the superior efficiency and performance of MetaLR compared to current SOTA medical fine-tuning techniques.

\section{Method}

This section provides a detailed description of the proposed MetaLR. It is a meta-learning-based~\cite{franceschi2018bilevel,li2017meta} approach that determines the appropriate LR for each layer based on its transfer feedback. It is important to note that fixing transferable layers is a special case of this method, where fixed layers always have zero LRs. First, we present the theoretical formulation of MetaLR. Next, we discuss online adaptation for efficiently determining optimal LRs. Finally, we demonstrate the use of a proportional hyper-LR and a validation scheme with batched training data to enhance performance.

\subsection{Formulation of Meta Learning Rate}\label{sect:formulation}

Let $(x,y)$ denotes a sample-label pair, and $\{(x_i,y_i): i=1, ..., N\}$ be the training data. The validation dataset $\{(x^v_i, y^v_i): i=1, ..., M\}$ is assumed to be independent and identically distributed as the training dataset. Let $\hat{y}=\Phi(x, \theta)$ be the prediction for sample $x$ from deep model $\Phi$ with parameters $\theta$.
In standard training of DNNs, the aim is to minimize the expected risk for the training set: $\frac{1}{N}\sum^{N}_{i=1}L(\hat{y}_i, y_i)$ with fixed training hyper-parameters, where $L(\hat{y}, y)$ is the loss function for the current task. The model generalization can be evaluated by the validation loss $\frac{1}{M}\sum^{M}_{i=1}L(\hat{y}^v_i, y^v_i)$. Based on the generalization, one can tune the hyper-parameters of the training process to improve the model.
The key idea of MetaLR is considering the layer-wise LRs as self-adaptive hyper-parameters during the training and automatically adjusting them to achieve better model generalization. We denote the LR and model parameters for the layer $j$ at the iteration $t$ as $\alpha_j^t$ and $\theta_j^t$. The LR scheduling scheme $\alpha = \{ \alpha_j^t: j=1, ..., d; ~t=1, ..., T\}$ is what MetaLR wants to learn, affecting which local optimal $\theta^*(\alpha)$ the model parameters $\theta^t = \{ \theta_j^t: j=1, ..., d\}$ will converge to. The optimal parameters $\theta^*(\alpha)$ are given by optimization on the training data. At the same time, the best LR tuning scheme $\alpha^*$ can be optimized based on the feedback for $\theta^*(\alpha)$ from the validation loss. This problem can be formulated as the following bi-level optimization problem:

\begin{equation}
\begin{aligned}
&\min_{\alpha} \frac{1}{M}\sum^{M}_{i=1}L(\Phi(x^v_i,\theta^*(\alpha)),y^v_i),\\
&s.t. ~\theta^*(\alpha)= \mathop{\arg\min}\limits_{\theta} \frac{1}{N}\sum^{N}_{i=1}L(\Phi(x_i,\theta),y_i).
\end{aligned}
\end{equation}

MetaLR aims to use the validation set to optimize $\alpha$ through an automatic process rather than a manual one. The optimal scheme $\alpha^*$ can be found by a nested optimization~\cite{franceschi2018bilevel}, but it is too computationally expensive in practice. A faster and more lightweight method is needed to make it practical.

\subsection{Onilne Learning Rate Adaptation}\label{sect:online}

Inspired by the online approximation \cite{ren2018learning}, we propose efficiently adapting the LRs and model parameters online.
The motivation of the online LR adaptation is updating the model parameters $\theta^t$ and LRs $\{\alpha^t_j:j=1,2,...,d\}$ within the same loop. We first inspect the descent direction of parameters $\theta^t_j$ on the training loss landscape and adjust the $\alpha^t_j$ based on the transfer feedback. Positive feedback (lower validation loss) means the LRs are encouraged to increase.

\begin{algorithm}[tb] 
\caption{Online Meta Learning Rate Algorithm} 
\label{algo} 
\begin{algorithmic}[1] 
\REQUIRE ~~\\ 
Training data $\mathcal{D}$, validation data $\mathcal{D}^v$, initial model parameter $\{\theta^0_1,...,\theta^0_d\}$, LRs $\{\alpha^0_1,...,\alpha^0_d\}$, batch size n, max iteration T;\\
\ENSURE ~~\\ 
Final model parameter $\theta^T=\{\theta^T_1,...,\theta^T_d\}$;
\FOR{$t = 0:T-1$}
\STATE $\{(x_i,y_i):i=1,...,n\}\leftarrow$ TrainDataLoader($\mathcal{D}$, n) ;\\
\STATE $\{(x^v_i,y^v_i):i=1,...,n\}\leftarrow$ ValidDataLoader($\mathcal{D}^v$, n) ;\\
\STATE Step forward for one step to get $\{\hat{\theta_1^t}(\alpha_1^t),..., \hat{\theta_d^t}(\alpha_d^t)\}$ with Eq. (\ref{eq:step1});\\
\STATE Update $\{\alpha^t_1,...,\alpha^t_d\}$ to become $\{\alpha^{t+1}_1,...,\alpha^{t+1}_d\}$ with Eq. (\ref{eq:step2});\\
\STATE Update $\{\theta^t_1,...,\theta^t_d\}$ to become $\{\theta^{t+1}_1,...,\theta^{t+1}_d\}$ with Eq. (\ref{eq:step3});\\
\ENDFOR
\end{algorithmic}
\end{algorithm}

We adopt Stochastic Gradient Descent (SGD) as the optimizer to conduct the meta-learning. The whole training process is summarized in Algorithm \ref{algo}. At the iteration $t$ of training, a training data batch $\{(x_i,y_i),i=1,...,n\}$ and a validation data batch $\{(x^v_i,y^v_i):i=1,...,n\}$ are sampled, where n is the size of the batches. First, the parameters of each layer are updated once with the current LR according to the descent direction on training batch.

\begin{equation}
\label{eq:step1}
    \hat{\theta_j^t}(\alpha_j^t) = \theta_j^t - \alpha_j^t \nabla_{\theta_j} (\frac{1}{n}\sum_{i=1}^n L(\Phi(x_i,\theta_j^t),y_i)),~j=1,...,d.
\end{equation}

This step of updating aims to get feedback for LR of each layer. After taking derivative of the validation loss \wrt $\alpha_j^t$, we can utilize the gradient to know how the LR for each layer should be adjusted. So the second step of MetaLR is to move the LRs along the meta objective gradient on the validation data:

\begin{equation}
\label{eq:step2}
    \alpha_j^{t+1} = \alpha_j^t - \eta \nabla_{\alpha_j} (\frac{1}{n}\sum_{i=1}^n L(\Phi(x_i^v,\hat{\theta_j^t}(\alpha_j^t)),y_i^v)),
\end{equation}

\noindent where $\eta$ is the hyper-LR. Finally, the updated LRs can be employed to optimize the model parameters through gradient descent truly.

\begin{equation}
\label{eq:step3}
    \theta_j^{t+1} = \theta_j^t - \alpha_j^{t+1} \nabla_{\theta_j} (\frac{1}{n}\sum_{i=1}^n L(\Phi(x_i,\theta_j^t),y_i)).
\end{equation}

For practical use, we constrain the LR for each layer to be $\alpha_j^{t} \in [10^{-6}, 10^{-2}]$.

\subsection{Proportional Hyper Learning Rate}

In practice, LRs are often tuned in an exponential style (\eg 1e-3, 3e-3, 1e-2) and are always positive values. However, if a constant hyper-LR is used, it will linearly update its corresponding LR regardless of numerical constraints. This can lead to fluctuations in the LR or even the risk of the LR becoming smaller than 0 and being truncated. To address this issue, we propose using a proportional hyper-LR $\eta=\beta \times \alpha_j^t$, where $\beta$ is a pre-defined hyper-parameter. This allows us to rewrite Eq. (\ref{eq:step2}) as:

\begin{equation}
\label{eq:new_step2}
    \alpha_j^{t+1} = \alpha_j^t (1  - \beta \nabla_{\alpha_j} (\frac{1}{n}\sum_{i=1}^n L(\Phi(x_i^v,\hat{\theta_j^t}(\alpha_j^t)),y_i^v))).
\end{equation}

The exponential update of $\alpha_j^t$ guarantees its numerical stability.

\subsection{Generalizability Validation on Training Data Batch}

One limitation of MetaLR is that the LRs are updated using separate validation data, which reduces the amount of data available for the training process. This can be particularly problematic for medical transfer learning, where the amount of downstream data has already been limited. In Eq. \ref{eq:step1} and Eq. \ref{eq:step2}, the update of model parameter $\theta_j^t$ and LR $\alpha_j^t$ is performed using different datasets to ensure that the updated $\theta_j^t$ can be evaluated for generalization without being influenced by the seen data. As an alternative, but weaker, approach, we explore using \textbf{another batch of training data} for Eq. \ref{eq:step2} to evaluate generalization. Since this batch was not used in the update of Eq. \ref{eq:step1}, it may still perform well for validation in meta-learning. The effect of this approach is verified in Sec. \ref{sect:ablation}, and the differences between the two methods are analyzed in Sec. \ref{sect:discussion}.

\section{Experiments and Analysis}

\subsection{Experimental Settings}

We extensively evaluate MetaLR on four transfer learning tasks (as shown in Tab. \ref{tab:tasks}).  To ensure the reproducibility of the results, all pre-trained models (USCL~\cite{chen2021uscl}, ImageNet~\cite{deng2009imagenet}, C2L~\cite{zhou2020comparing}, Models Genesis~\cite{zhou2019models}) and target datasets (POCUS~\cite{born2020pocovid}, BUSI~\cite{al2020dataset}, Chest X-ray~\cite{kermany2018large}, LiTS~\cite{bilic2019liver}) are publicly available. In our work, we consider models pre-trained on both natural and medical image datasets, with three target modalities and three target organs, which makes our experimental results more credible. For the lesion detection tasks, we used ResNet-18~\cite{he2016deep} with the Adam optimizer. The initial learning rate (LR) and hyper-LR coefficient $\beta$ are set to $10^{-3}$ and $0.1$, respectively. In addition, we use 25\% of the training set as the validation set for meta-learning. For the segmentation task, we use 3D U-Net~\cite{cciccek20163d} with the SGD optimizer. The initial LR and hyper-LR coefficient $\beta$ are set to $10^{-2}$ and $3\times10^{-3}$, respectively. The validation set for the LiTS segmentation dataset comprises 23 samples from the training set of size 111. All experiments are implemented using PyTorch 1.10 on an Nvidia RTX A6000 GPU. We report the mean values and standard deviations for each experiment with five different random seeds. For more detailed information on the models and hyper-parameters, please refer to our supplementary material.

\begin{table}[t]
\tiny
\centering
\caption{Pre-training data, algorithms, and target tasks.}
\setlength{\tabcolsep}{1.0mm} 
\renewcommand{\arraystretch}{1.2} 
\begin{tabular}{cc|ccccc}
\toprule
\rule{0pt}{4pt}
Source & Pre-train Method  & Target & Object & Task & Modality & Size        \\
\hline
\rule{0pt}{6pt}
US-4~\cite{chen2021uscl} & USCL~\cite{chen2021uscl} & POCUS~\cite{born2020pocovid} & Lung & COVID-19 detection & US & 2116 images \\
\rule{0pt}{6pt}
ImageNet~\cite{deng2009imagenet} & supervised & BUSI~\cite{al2020dataset} & Breast & Tumor detection & US & 780 images  \\
\rule{0pt}{6pt}
MIMIC-CXR~\cite{johnson2019mimic} & C2L~\cite{zhou2020comparing} & Chest X-ray~\cite{kermany2018large} & Lung & Pneumonia detection & X-ray & 5856 images \\
\rule{0pt}{6pt}
LIDC-IDRI~\cite{armato2011lung} & Models Genesis~\cite{zhou2019models} & LiTS~\cite{bilic2019liver} & Liver & Liver segmentation & CT & 131 volumes\\
\bottomrule
\end{tabular}
\label{tab:tasks}
\end{table}

\subsection{Ablation Study}\label{sect:ablation}

In order to evaluate the effectiveness of our proposed method, we conduct an ablation study \wrt the basic MetaLR algorithm, the proportional hyper-LR, and batched-training-data validation (as shown in Tab. \ref{tab:ablation}). When applying only the basic MetaLR, we observe only marginal performance improvements for the four downstream tasks. We conjecture that this is due to two reasons: Firstly, the constant hyper-LR makes the training procedures less stable than direct training, which is evident from the larger standard deviation of performance. Secondly, part of the training data are split for validation, which can be detrimental to the performance. After applying the proportional hyper-LR, significant improvements are in both the performance and its stability. Moreover, although the generalization validation on the training data batch may introduce bias, providing sufficient training data ultimately benefits the performance.

\begin{table}[t]
\scriptsize
\centering
\caption{Ablation study for MetaLR, hyper-LR, and validation data. The baseline is the direct tuning of all layers with constant LRs. The default setting for MetaLR is a constant hyper-LR of $10^{-3}$ and a separate validation set.}
\setlength{\tabcolsep}{1.0mm} 
\renewcommand{\arraystretch}{1.2} 
\begin{tabular}{ccc|c|c|c|c}
\toprule
\rule{0pt}{4pt}
MetaLR & Prop. hyper-LR & Val. on trainset  & POCUS & BUSI & Chest X-ray & LiTS        \\
\hline
\rule{0pt}{6pt}
& &  & 91.6$\pm$0.8 & 84.4$\pm$0.7 & 94.8$\pm$0.3 & 93.1$\pm$0.4 \\
\checkmark & &  & 91.9$\pm$0.6 & 84.9$\pm$1.3 & 95.0$\pm$0.4 & 93.2$\pm$0.8 \\
\checkmark & \checkmark &  & 93.6$\pm$0.4 & 85.2$\pm$0.8 & 95.3$\pm$0.2 & 93.3$\pm$0.6 \\
\checkmark & & \checkmark & 93.0$\pm$0.3 & 86.3$\pm$0.7 & 95.5$\pm$0.2 & 93.9$\pm$0.5 \\
\checkmark & \checkmark & \checkmark & \textbf{93.9}$\pm$0.4 & \textbf{86.7}$\pm$0.7 & \textbf{95.8}$\pm$0.3 & \textbf{94.2}$\pm$0.5 \\
\bottomrule
\multicolumn{7}{l}{\scriptsize $\bullet$ Final MetaLR outperforms baseline with p-values of $0.0014, 0.0016, 0.0013, 0.0054$.}\\
\end{tabular}
\label{tab:ablation}
\vspace{-0.3cm}    
\end{table}

\subsection{Comparative Experiments}

In our study, we compare MetaLR with several other fine-tuning schemes, including tuning only the last layer / all layers with constant LRs, layer-wise fine-tuning~\cite{tajbakhsh2016convolutional}, bi-directional fine-tuning~\cite{chen2022rethinking}, and AutoLR~\cite{ro2021autolr}. The U-Net fine-tuning scheme proposed by Mina \etal~\cite{amiri2020fine} was also evaluated.

\begin{table}[t]
\tiny
\centering
\caption{Comparative experiments on lesion detection. We report sensitivities (\%) of the abnormalities, overall accuracy (\%), and training time on each task.}
\setlength{\tabcolsep}{0.0mm} 
\renewcommand{\arraystretch}{1.5} 
\begin{tabular}{c|c|c|c|c|c|c|c|c|c|c|c}
\toprule
\multicolumn{1}{c|}{\multirow{2}{*}{Method}} &  \multicolumn{4}{c|}{POCUS} & \multicolumn{4}{c|}{BUSI} & \multicolumn{3}{c}{Chest X-ray} \\
\multicolumn{1}{c|}{} & ~~COVID~~ & ~~~Pneu.~~~ & ~~~~Acc~~~~ & ~Time~ & ~~Benign~~ & ~Malignant~ & ~~~~Acc~~~~ & ~Time~ & ~~~Pneu.~~~ & ~~~~Acc~~~~ & ~Time~ \\
\hline
\hline
\rule{0pt}{4pt} 
Last Layer & 77.9$\pm$2.1 & 84.0$\pm$1.3 & 84.1$\pm$0.2 & 15.8m & 83.5$\pm$0.4 & 47.6$\pm$4.4 & 66.8$\pm$0.5 & 4.4m & 99.7$\pm$1.3 & 87.8$\pm$0.6 & 12.7m \\
All Layers & 85.8$\pm$1.7 & 90.0$\pm$1.9 & 91.6$\pm$0.8 & 16.0m & 90.4$\pm$1.5 & 77.8$\pm$3.5 & 84.4$\pm$0.7 & 4.3m & 98.8$\pm$0.2 & 94.8$\pm$0.3 & 12.9m \\
Layer-wise & 87.5$\pm$1.0 & 92.3$\pm$1.3 & 92.1$\pm$0.3 & 2.4h & 90.8$\pm$1.2 & 75.7$\pm$2.6 & 85.6$\pm$0.4 & 39.0m &  97.9$\pm$0.3 & 95.2$\pm$0.2 & 1.9h \\
Bi-direc. & 90.1$\pm$1.2 & 92.5$\pm$1.5 & 93.6$\pm$0.2 & 12.0h & 92.2$\pm$1.0 & 77.1$\pm$3.5 & 86.5$\pm$0.5 & 3.2h & 98.4$\pm$0.3 & 95.4$\pm$0.1 & 9.7h \\
AutoLR & 89.8$\pm$1.6 & 89.7$\pm$1.5 & 90.4$\pm$0.8 & 17.5m & 90.4$\pm$1.8 & 76.2$\pm$3.2 & 84.9$\pm$0.8 & 4.9m & 95.4$\pm$0.5 & 93.0$\pm$0.8 & 13.3m \\
\hline
MetaLR & 94.8$\pm$1.2 & 93.1$\pm$1.5 &  \textbf{93.9}$\pm$0.4 & 24.8m & 92.2$\pm$0.7 & 75.6$\pm$3.6 & \textbf{86.7}$\pm$0.7 & 6.0m & 97.4$\pm$0.4 & \textbf{95.8}$\pm$0.3 & 26.3m \\
\bottomrule
\end{tabular}
\label{tab:comparison_classification}
\vspace{-0.3cm}
\end{table}

\textbf{Results on Lesion Detection Tasks.} MetaLR consistently shows the best performance on all downstream tasks (Tab. \ref{tab:comparison_classification}). It shows 1\% - 2.3\% accuracy improvements compared to direct training (\ie tuning all layers) because it takes into account the different transferabilities of different layers. While manual picking methods, such as layer-wise and bi-directional fine-tuning, also achieve higher performance, they require much more training time (5$\times$ - 50$\times$) for searching the best tuning scheme. On the other hand, AutoLR is efficient, but its strong hypothesis harms its performance sometimes. In contrast, MetaLR makes no hypothesis about transferability and learns appropriate layer-wise LRs on different domains. Moreover, its performance improvements are gained with only 1.5$\times$ - 2.5$\times$ training time compared with direct training.

\begin{table}[t]
\scriptsize
\centering
\caption{Comparative experiments on LiTS liver segmentation task.}
\setlength{\tabcolsep}{2mm} 
\renewcommand{\arraystretch}{1.2} 
\begin{tabular}{c|c|c|c|c}
\toprule
Method & ~~~PPV~~~ & ~Sensitivity~ & ~~~Dice~~~ & ~Time~ \\
\hline
\hline
\rule{0pt}{8pt} 
Last Layer & 26.1$\pm$5.5 &  71.5$\pm$4.2 &  33.5$\pm$3.4 & 2.5h  \\
All Layers & 94.0$\pm$0.6 &  93.1$\pm$0.7 & 93.1$\pm$0.4 & 2.6h  \\
Layer-wise & 92.1$\pm$1.3 &  96.4$\pm$0.4 & 93.7$\pm$0.3 & 41.6h \\
Bi-direc. & 92.4$\pm$1.1 & 96.1$\pm$0.2 & 93.8$\pm$0.1 & 171.2h \\
Mina \etal & 92.7$\pm$1.2 & 93.2$\pm$0.5 & 92.4$\pm$0.5 & 2.6h  \\
\hline
MetaLR & 94.4$\pm$0.9  &  93.6$\pm$0.4 &  \textbf{94.2}$\pm$0.5 & 5.8h \\
\bottomrule
\end{tabular}
\label{tab:comparison_segmentation}
\end{table}

\textbf{Results on Segmentation Task.} MetaLR achieves the best Dice performance on the LiTS segmentation task (Tab. \ref{tab:comparison_segmentation}). Unlike ResNet for lesion detection, the U-Net family has a more complex network topology. With skip connections, there are two interpretations~\cite{amiri2020fine} of depths for layers: 1) the left-most layers are the shallowest, and 2) the top layers of the ``U” are the shallowest. This makes the handpicking methods even more computationally expensive. However, MetaLR updates the LR for each layer according to their validation gradients, and its training efficiency is not affected by the complex model architecture.

\subsection{Discussion and Findings}\label{sect:discussion}

\begin{figure}[t]
\vspace{0cm}                          
\centering\centerline{\includegraphics[width=1.0\linewidth]{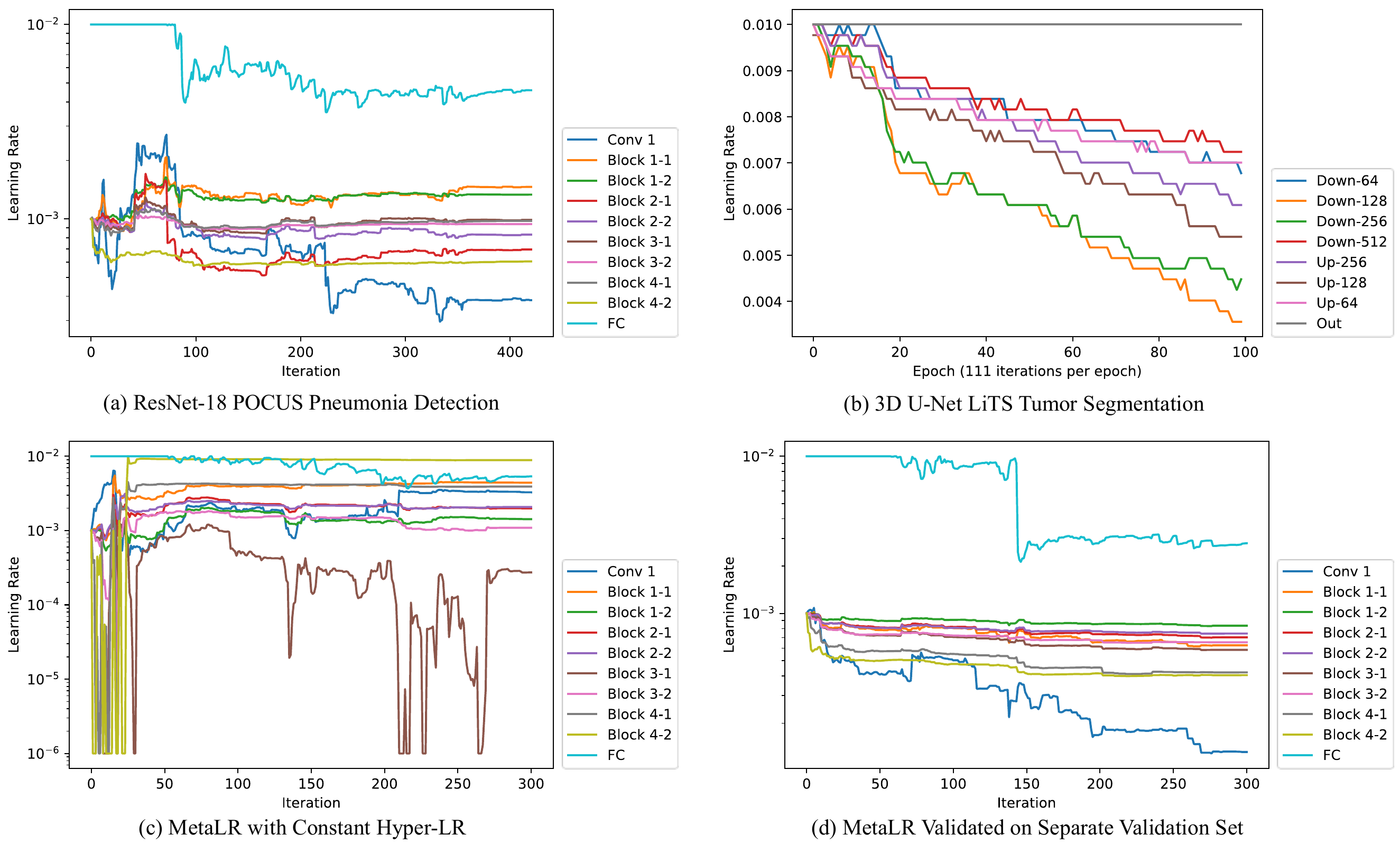}}
\caption{The LR curves for MetaLR on POCUS detection (a), on LiTS segmentation (b), with constant hyper-LR (c), and with a separate validation set (d).}
\label{fig:discussion}
\end{figure}

The LRs learned by MetaLR change as the training progresses. For ResNet-18 (Fig. \ref{fig:discussion} (a)), the layer-wise LRs fluctuate drastically during the first 100 iterations. However, after iteration 100, all layers except the first layer ``Conv1” become stable at different levels. The first layer has a decreasing LR  (from $2.8\times 10^{-3}$ to $3\times10^{-4}$) throughout the process, reflecting its higher transferability. For 3D U-Net (Fig. \ref{fig:discussion} (b)), the middle layers of the encoder ``Down-128” and ``Down-256” are the most transferable and have the lowest LRs, which is difficult for previous fine-tuning schemes to discover. As expected, the randomly initialized ``FC” and ``Out” layers have the largest LRs since they are not transferable. 

We also illustrate the LR curves with a constant hyper-LR instead of a proportional one. The LR curves of ``Block 3-1” and ``Block 4-2” become much more fluctuated (Fig. \ref{fig:discussion} (c)). This instability may be the key reason for the instability of performance when using a constant hyper-LR. Furthermore, we surprisingly find that the learned LRs are similar to the curves learned when validated on the training set when using a separate validation set Fig. \ref{fig:discussion} (d)). With similar learned LR curves and more training data, it is reasonable that batched training set validation can be an effective alternative to the basic MetaLR.

\section{Conclusion}

In this work, we proposed a new fine-tuning scheme, MetaLR, for medical transfer learning. It achieves significantly superior performance to the previous SOTA fine-tuning algorithms. MetaLR alternatively optimizes model parameters and layer-wise LRs in an online meta-learning fashion with a proportional hyper-LR. It learns to assign lower LRs for the layers with higher transferability and higher LRs for the less transferable layers. The proposed algorithm is easy to implement and shows the potential to replace manual layer-wise fine-tuning schemes. Future works include adapting MetaLR to a wider variety of clinical tasks.

\section{Acknowledgement}
This work was supported by the National Natural Science Foundation of
China (No. 62101351) and the GuangDong Basic and Applied Basic Research Foundation
(No.2020A1515110376).

\bibliographystyle{splncs04}
\bibliography{MetaLR}

\end{document}